\documentclass[11pt]{article}

\usepackage[final]{acl}

\usepackage{times}
\usepackage{latexsym}

\usepackage[T1]{fontenc}
\usepackage[utf8]{inputenc}

\usepackage{microtype}

\usepackage{inconsolata}

\usepackage{graphicx}

\usepackage{graphicx}
\usepackage{booktabs}
\usepackage{multirow}
\usepackage{bbm}
\usepackage{listings}
\usepackage[normalem]{ulem} % provides \uline{}
\usepackage{enumitem}
\usepackage{subcaption}
\usepackage[table]{xcolor} 
\usepackage{xcolor}
\usepackage{amsmath}
\usepackage{pifont}
\usepackage{array}
\usepackage{caption}
\usepackage{amssymb}

\usepackage{tcolorbox}
\tcbuselibrary{skins}
\usepackage[linesnumbered,ruled,vlined]{algorithm2e}

\definecolor{best}{HTML}{fd8d3c}  % 深橙色
\definecolor{better}{HTML}{feb24c} % 中橙色
\definecolor{good}{HTML}{ffeda0}   % 浅橙色
\definecolor{headergray}{HTML}{e9e9e9} % 标题灰色
\definecolor{promptborder}{HTML}{7E4A9E} % 深紫色边框/页眉
\definecolor{promptbg}{HTML}{F6F1FA}     % 浅薰衣草色背景
\definecolor{judgeframe}{HTML}{CCCCCC}    % 极细灰色边框 (#CCCCCC)
\definecolor{judgeheader}{HTML}{EBEBEB}   % 标题栏浅灰背景 (#EBEBEB)
\definecolor{judgebody}{HTML}{F9F9F9}     % 内容区近乎白色的极浅灰 (#F9F9F9)
\newcommand{\chk}{\ensuremath{^{\textcolor[HTML]{1E8449}{\mathbf{\checkmark}}}}}
\newcommand{\cls}{\ensuremath{^{\textcolor[HTML]{C0392B}{\mathbf{\times}}}}}

\newcolumntype{C}{>{\centering\arraybackslash}p{1.8cm}}
\newcolumntype{L}{>{\raggedright\arraybackslash}p{3.2cm}}
\newtcolorbox{judgepromptbox}[1]{
  colback=judgebody,
  colframe=judgeframe,
  colbacktitle=judgeheader,
  coltitle=black,
  fonttitle=\bfseries\sffamily\small,
  title={#1},
  arc=1.2mm,               % 极小的圆角
  boxrule=0.6pt,           % 细边框
  left=12pt, right=12pt, top=12pt, bottom=12pt,
  toptitle=5pt, bottomtitle=5pt,
  enhanced,
  attach boxed title to top left={yshift=-0mm, xshift=0mm}, 
  segmentation style={draw=judgeframe, line width=0.6pt}
}

\definecolor{tblgray}{gray}{0.94} % 定义一种浅灰色

\title{QueryAgent-R1: Bridging Query Generation and Product Retrieval for E-Commerce Query Recommendation}

\author{Dike Sun$^{1}$, Zheng Zou$^{1}$, Jingtong Zang$^{1}$, Qi Sun$^{1}$\thanks{Corresponding author.}, Huaipeng Zhao$^{1}$,Tao Luo $^{1}$,  
    Xiaoyi Zeng $^{1}$  \\
  $^{1}$Alibaba International Digital Commercial Group \\ 
  \texttt{\{sundike.sdk,qiran.sq\}@alibaba-inc.com}, \\
  }

\begin{document}
\maketitle
\begin{abstract}

Query recommendation in e-commerce search aims to proactively suggest queries that match users' potential interests. However, existing methods mainly optimize query-level relevance, while neglecting whether the retrieved products align with users' downstream preferences. This mismatch often leads to high query click through rates (CTR) but low product conversion rates (CVR). To bridge this gap, we propose \textbf{QueryAgent-R1}, a memory-augmented agentic framework that improves end-to-end alignment via chain-of-retrieval optimization. Our QueryAgent-R1 grounds query generation in real inventory retrieval, allowing the agent to validate and refine queries based on retrieved products. We also design a consistency reward in the agentic reinforcement learning (RL) process to jointly optimize query relevance and downstream engagement. In addition, we construct a memory abstraction module for efficient user profiling. To support offline evaluation, we construct two datasets based on both proprietary industrial data and public datasets, on which QueryAgent-R1 consistently outperforms strong baselines. Moreover, on a large scale production platform, QueryAgent-R1 improves Query CTR by 2.9\% and guided CVR by 3.1\% in online A/B tests. 

\end{abstract}

\section{Introduction}

\begin{figure}
	\centering
	\includegraphics[width=0.3\textheight]{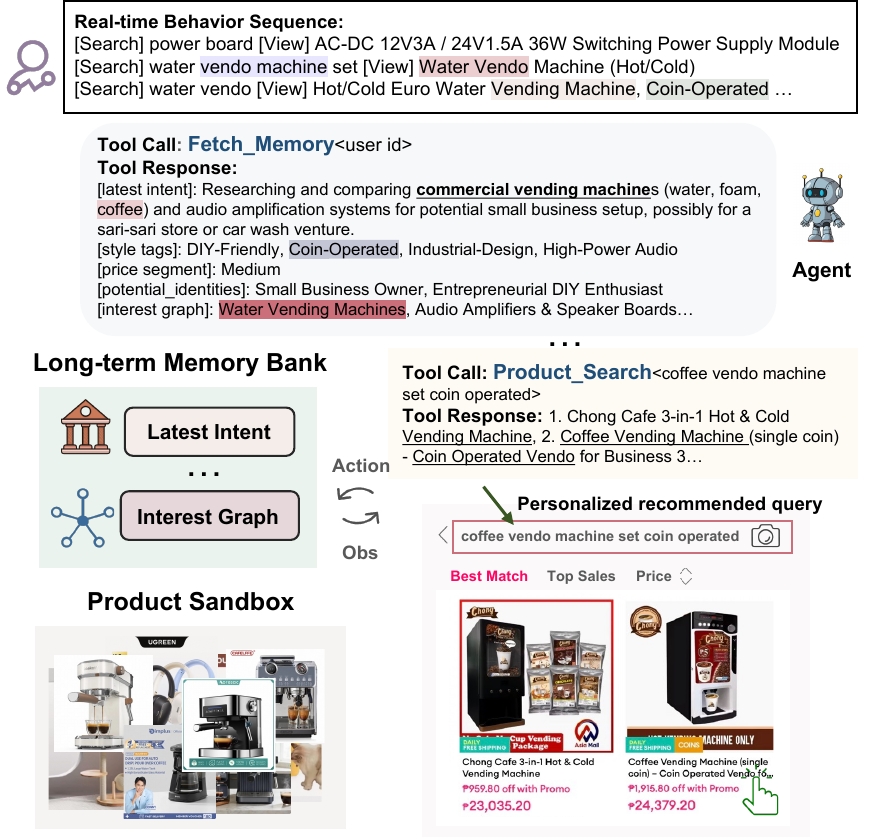}
    
	\caption{Illustration of the query recommendation process in QueryAgent-R1. Without a current query context, the system generates the personalized query from the user's historical behavior sequence.}
    \vspace{-0.9em}
	\label{fig:intro}  
 
\end{figure}

Query recommendation~\cite{query_rec04,lai2023workload,bacciu2024generating,min2025promoting,guo2025onesug} plays a crucial role in e-commerce search systems by proactively suggesting queries that match users' potential interests and guiding them toward relevant products. Especially in the search-bar placeholder setting~\cite{zeus2021,aigq}, where only a single pre-filled query is shown before any user input (Figure \ref{fig:intro}), recommendation quality directly affects both user engagement and downstream conversion.

Existing industrial methods rely heavily on inventory-based matching (e.g., ItemCF~\cite{itemcf}, Swing~\cite{swing}) or independent semantic retrieval (e.g., DSSM~\cite{dssm}, LLM-based models~\cite{qwen3_emb}). While efficient, the above methods primarily optimize query-level relevance, i.e., whether a recommended query appears attractive or semantically aligned with user interests. However, they often overlook a more important question: whether the products retrieved by that query are actually consistent with the user's downstream preferences and likely to induce further engagement. This mismatch creates a \emph{generation-retrieval gap}: a recommended query may attract clicks and achieve high click-through rate (CTR), while the retrieved products fail to satisfy purchase intent, resulting in poor post-click engagement and low product conversion. 

To bridge this gap, we propose \textbf{QueryAgent-R1}, a memory-augmented agentic framework for query recommendation in e-commerce search. Specifically, we ground query generation in real inventory retrieval, allowing our agent to validate the generated query through retrieved products and refinement. In addition, we propose a consistency reward for agentic RL process to jointly optimize query relevance and downstream engagement. We also design a memory abstraction mechanism that extracts an interest graph from users' long-term memory for efficient user profiling.

In addition, we construct two datasets based on proprietary industrial and public data for offline evaluation, and further deploy QueryAgent-R1 on a large-scale e-commerce platform with tens of millions of daily active users. Experimental results show that QueryAgent-R1 consistently outperforms strong baselines on the offline datasets and delivers significant gains in online A/B tests, improving Query CTR by 2.9\% and guided CVR by 3.1\%. Our main contributions are summarized as follows:
% [leftmargin=*, itemsep=2pt, topsep=2pt, parsep=0pt]
\begin{itemize}[noitemsep]
     \item We propose QueryAgent-R1, a memory-augmented agentic framework for query recommendation combining retrieval-grounded generation with a consistency reward for RL alignment.
    \item We propose a memory abstraction mechanism that extracts an interest graph from users' long-term memory for efficient user profiling.
    \item We construct two datasets from industrial and public data for offline evaluation. Experimental results show that our QueryAgent-R1 achieves significant improvements in both offline and online settings.
\end{itemize}

\begin{figure}
  \centering
 \includegraphics[width=0.30\textheight]{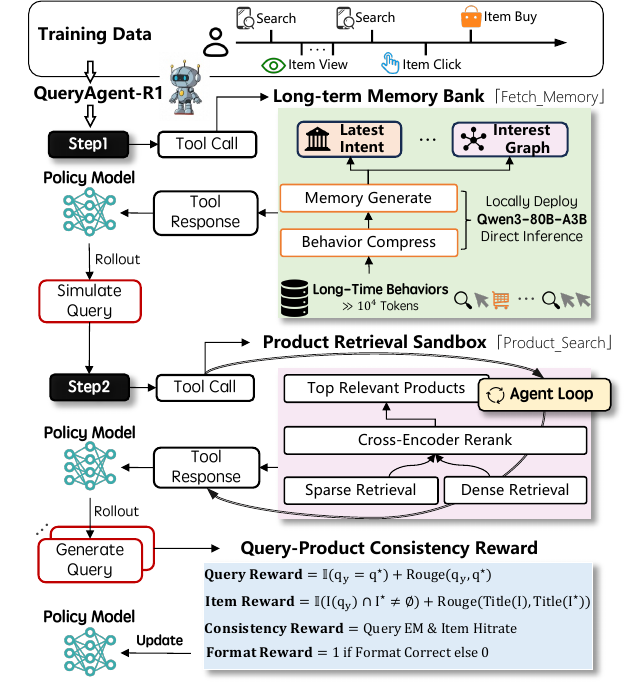}
  \caption{%
    \textbf{QueryAgent-R1 Architecture.} The policy model is optimized via RL to interleave a Memory Tool (compressing long logs) and a Search Tool (grounding generation in real inventory), unified by a Chain-of-Retrieval outcome reward.%
  }
  \vspace{-0.9em}
  \label{fig:architecture}
\end{figure}

\section{Problem Formulation}
Let $\mathcal{U}$ and $\mathcal{I}$ denote the sets of users and items. For each user $u \in \mathcal{U}$, we consider a real-time interaction sequence $\mathcal{H}_u = \{(a_i, c_i)\}_{i=1}^{N}$ of the $N{=}50$ most recent interactions, where $a_i \in \{\texttt{search}, \texttt{click}\}$ is the action type and $c_i$ is the textual context.

We formulate search-bar placeholder generation as a Chain-of-Retrieval optimization problem. Our objective is to find a query $q_y$ that maximizes the joint probability of query click and subsequent item click:
\begin{equation}
    \operatorname*{max}_{q_y \in \mathcal{V}} \underbrace{P(q_y=q^{\star} \mid  \mathcal{H}_u)}_{\text{Hop-1: Query CTR}} \cdot \underbrace{P(\mathcal{I}(q_y)\cap \mathcal{I}^{\star}\neq\emptyset \mid \mathcal{H}_u)}_{\text{Hop-2: Item CTR}},
\end{equation}
% \begin{equation}
%     \operatorname*{max}_{q_y \in \mathcal{V}} \underbrace{P(z_{q_y}=1 \mid  \mathcal{H}_u)}_{\text{Hop-1: Query CTR}} \cdot \underbrace{P(z_{\mathcal{I}(q_y)}=1 \mid \mathcal{H}_u)}_{\text{Hop-2: Item CTR}},
% \end{equation}
where $\mathcal{V}$ denotes the query vocabulary and $\mathcal{I}(q_y)$ denotes the items retrieved by query $q_y$. Here, $q^{\star}$ and $\mathcal{I}^{\star}$ denote the ground-truth query and retrieved item set, respectively. In practice, we use item clicks as a dense proxy for sparse conversion signals.

\section{Methodology}
As shown in Figure~\ref{fig:architecture}, our \textbf{QueryAgent-R1}, built on Qwen3-4B, operates in memory and product sandboxes with reinforcement learning guided by a collaborative reward that jointly optimizes query and product preferences.

\subsection{Memory Abstraction Mechanism}
We design a memory environment with dedicated tools for memory storage, compression, and retrieval based on users' ultra-long behavioral histories, which encompass all user behaviors over a three-month period. To avoid injecting extensive online behavior logs ($>>$10$^4$ tokens) into the context, we use Qwen3-Next-80B-A3B~\cite{yang2025qwen3technicalreport} to asynchronously compress raw user behaviors offline into a compact user profile (prompt in Appendix~\ref{fig:memory_abstract_prompt}). When our agent executes a \textit{Fetch\_Memory} action, it directly retrieves the cached profile (Figure~\ref{fig:intro}), which includes the latest intent, interest graph, potential identities, and other personalized signals. By incrementally refreshing the profile, we achieve 10$\times$ sequence compression while preserving up-to-date personalization.

\subsection{Product Retrieval Augmentation}
We build a product retrieval sandbox with four million real-world products, filtered from a live e-commerce environment, to ground the policy in real products and reduce the gap between generated queries and retrieved products. This allows the agent to verify whether a generated query can retrieve relevant and purchasable products before the final recommendation.
 The retrieval tool employs a hybrid strategy integrating BM25~\cite{bm25} (sparse retrieval) and Qwen3-Embedding-0.6B~\cite{qwen3_emb} (dense retrieval), followed by cross-encoder reranking via Qwen3-Rerank-0.6B~\cite{qwen3_emb}. During query generation, our agent calls this tool to inspect the products returned by the query and refine the query accordingly. This helps the agent generate queries that are not only attractive, but also effective at retrieving relevant and purchasable products.

\subsection{Agentic RL Training}
We further optimize the policy model with reinforcement learning. Given a complete agent trajectory $y$, we define the reward as
\begin{equation}
R(y) =
\lambda_{\mathrm{fmt}} r_{\mathrm{fmt}}(y)
+ \lambda_{\mathrm{tool}} r_{\mathrm{tool}}(y)
+ \lambda_{\mathrm{hit}} r_{\mathrm{hit}}(y).
\end{equation}
Here, $r_{\mathrm{fmt}}(y)$ evaluates whether the trajectory follows the required structured format, $r_{\mathrm{tool}}(y)$ rewards correct invocation of the memory and product retrieval tools, and $r_{\mathrm{hit}}(y)$ measures whether the generated query leads to the desired retrieval product. We further define
\begin{equation}
r_{\mathrm{hit}}(y)
=
\lambda_q r_q(y)
+
\lambda_i r_i(y),
\end{equation}
To mitigate reward sparsity, we decompose $r_q(y)$ and $r_i(y)$ into hard and soft components. The query reward is defined as $r_q(y) = \mathbb{I}(q_y = q^\star) + \mathrm{ROUGE}(q_y, q^\star)$, where query-level ROUGE~\cite{lin2003rouge} provides dense supervision to bootstrap the sparse Exact Match~\cite{yu2024rankragunifyingcontextranking}. Similarly, the item reward $r_i(y) = \mathbb{I}\bigl(\mathcal{I}(q_y) \cap \mathcal{I}^\star \neq \emptyset\bigr) + \mathrm{ROUGE}\bigl(\mathrm{Title}(I), \mathrm{Title}(I^\star)\bigr)$ combines the sparse HitRate~\cite{hitrate} with title-level ROUGE to facilitate policy exploration. Here, $\mathbb{I}(\cdot)$ is the indicator function, $\mathcal{I}(q_y)$ and $\mathcal{I}^\star$ denote the items retrieved by $q_y$ and the ground-truth clicked items from offline logs, respectively.

% where $r_q(y)$ is the ROUGE similarity~\cite{lin2003rouge} between the generated query and the reference query from offline logs, and $r_i(y)$ indicates whether the retrieved item set contains the ground-truth clicked item. Formally, let $\mathcal{I}(q_y)$ denote the set of items retrieved by the generated query $q_y$, $\mathcal{I}^\star$ denote the ground-truth clicked item set from offline logs, and $\mathbb{I}(\cdot)$ denote the indicator function:

% \begin{equation}
% r_q(y) = \mathrm{ROUGE}(q_y, q^\star),
% \end{equation}
% \begin{equation}
% r_i(y)
% =
% \mathbb{I}
% \bigl(
% \mathcal{I}(q_y) \cap \mathcal{I}^\star \neq \emptyset
% \bigr).
% \end{equation}

We optimize the policy using GDPO~\cite{gdpo}. For each input $x$, we rollout $K$ trajectories
$G = \{y_1, \dots, y_K\}$ and update the policy based on group relative advantages. Compared with GRPO~\cite{grpo}, GDPO uses decoupled normalization over reward components, which better fits our reward design with both dense and sparse signals. Detailed RL comparisons are in Appendix~\ref{appendix:rl_comparison}, and training prompts are in Appendix~\ref{appendix:training_prompt}.

\section{Experiments}
\label{sec:experiments}

\subsection{Experimental Setup}
\label{sec:setup}

\paragraph{Datasets.} We evaluate on two large-scale datasets:
1)~\textbf{Industrial}: 54k active users ($\geq$30 events/week) from a major e-commerce platform for training, and 5k for testing.
2)~\textbf{Public}: We merge Amazon ESCI~\cite{amazon_esci} (Apache-2.0 License) and Amazon Review Data~\cite{amazon_review} (MIT License) via product IDs (16k train / 1k test, Details in Appendix~\ref{app:dataset_construction}). 
For both, the $N{=}50$ most recent events act as context. Ground truth is hierarchical: a query is \emph{Hop-1 positive} if searched/clicked; an item is \emph{Hop-2 positive} if engaged post-search. 

\paragraph{Implementation.} We train Qwen3-4B as the backbone on \(8\times\) H20 GPUs using GDPO (\(\text{rollout}=8\), \(\text{lr}=10^{-5}\)). Each training run takes approximately 70 hours. Appendix~\ref{appendix:implementation_details}, ~\ref{appendix:sandbox} for details.

\paragraph{Baselines.} We mainly compare against two categories of baselines: 1) \textbf{inventory-based retrieval} methods, including Swing~\cite{swing}, a graph-based collaborative filtering method, and Qwen3-Emb-0.6B / 4B~\cite{qwen3_emb}, which retrieve historical queries via dense embedding similarity; and 2) \textbf{LLM direct inference} methods, including proprietary LLMs, directly generate recommendation queries from raw user behavioral logs.

\paragraph{Metrics.}Following previous work~\cite{yu2024rankragunifyingcontextranking, search-r1,deepretrieval,r1-searcher,r1-searcher++,convsearch-r1}, we use Q\_EM, which measures exact-match query generation, I\_Hit@$1$, measuring item retrieval success, and Cons@$1$, which requires both to hold simultaneously.

\subsection{Main Experiments}
\paragraph{Offline Results}

\begin{table}[t]
\centering

\resizebox{\columnwidth}{!}{%
\begin{tabular}{l ccc ccc}
\toprule
 & \multicolumn{3}{c}{\textbf{Industrial}}
 & \multicolumn{3}{c}{\textbf{Amazon}} \\
\cmidrule(lr){2-4} \cmidrule(lr){5-7}
\textbf{Method}
  & \textbf{Q\_EM} & \textbf{I\_Hit@1} & \textbf{Cons@1}
  & \textbf{Q\_EM} & \textbf{I\_Hit@1} & \textbf{Cons@1} \\
\midrule
\rowcolor{tblgray}
\multicolumn{7}{l}{\textit{Inventory-based Retrieval}} \\
Swing
  & 0.013 & 0.060 & 0.011
  & 0.037 & 0.006 & 0.003 \\
Qwen3-Emb-0.6B
  & 0.046 & 0.102 & 0.032
  & 0.033 & 0.008 & 0.003 \\
Qwen3-Emb-4B
  & 0.057 & 0.114 & 0.041
  & 0.041 & 0.042 & 0.002 \\
\rowcolor{tblgray}
\multicolumn{7}{l}{\textit{LLM Direct Inference}} \\
Qwen3.6-plus 
&0.047 &0.084 &0.028 
&0.096 &0.058 &0.015 \\
Qwen3-Max
  & 0.034 & 0.060 & 0.021
  & 0.035 & 0.047 & 0.001 \\
Gemini-3.1-pro 
& 0.054 & 0.096 & 0.029 
& \textbf{0.123} & 0.091 & 0.021 \\
Gemini-3.0-pro
  & 0.040 & 0.065 & 0.018
  & 0.115 & 0.080 & 0.019\\
GPT-5.1
  & 0.038 & 0.091 & 0.023
  & 0.039 & 0.042 & 0.002 \\
GPT-4o      
  & 0.036 & 0.065 & 0.020 
  & 0.033 & 0.030 & 0.002 \\

DeepSeek-v4-flash   
& 0.049 & 0.064 & 0.023 
& 0.070 & 0.091 & 0.013 \\

DeepSeek-v4-pro     
& 0.034 & 0.070 & 0.018 
& 0.082 & 0.091 & 0.011 \\

\rowcolor{tblgray}
\multicolumn{7}{l}{\textit{Ours}} \\
Qwen3-4B Backbone
  & 0.037 & 0.084 & 0.025
  & 0.041 & 0.042 & 0.002 \\
+Retrieval
  & 0.023 & 0.091 & 0.018
  & 0.083 & 0.077 & 0.013 \\
+RL training
  & 0.032 & 0.102 & 0.021
  & 0.057 & 0.062 & 0.009 \\
\midrule
\textbf{QueryAgent-R1}
  & \textbf{0.177} & \textbf{0.261} & \textbf{0.117}
  & 0.095 & \textbf{0.144} & \textbf{0.063} \\
\bottomrule
\end{tabular}%
}
\caption{%
  Offline experimental results on industrial and public datasets.}
\vspace{-0.9em}
\label{tab:offline}

% \vspace{-8mm}
\end{table}
As shown in Table~\ref{tab:offline}, our QueryAgent-R1 achieves the strongest end-to-end performance on both datasets, with particularly large gains on Cons@1. On Industrial, it reaches 0.177 Q\_EM, 0.261 I\_Hit@1, and 0.117 Cons@1, clearly outperforming the strongest inventory-based baseline Qwen3-Emb-4B and the strongest direct-inference LLM baseline (Details in Appendix~\ref{appendix:LLM_direct_gap_analysis}) Gemini-3.1-pro. On Amazon, although Gemini-3.1-pro obtains higher Q\_EM, QueryAgent-R1 achieves the best I\_Hit@1 (0.144) and Cons@1 (0.063), far exceeding their Cons@1 scores. Compared with the Qwen3-4B backbone, QueryAgent-R1 improves Cons@1 from 0.025 to 0.117 on Industrial and from 0.002 to 0.063 on Amazon, confirming the effectiveness of our end-to-end RL optimization. Moreover, simply adding retrieval or RL brings limited gains, while the strong improvements of QueryAgent-R1 show that our collaborative reward better aligns query generation with downstream retrieval. 

\paragraph{Online A/B Test.} QueryAgent-R1 was deployed on $1\%$ of live traffic for a cumulative $7$-day online A/B test, serving millions of requests. Compared to the production baseline, it achieved a $2.9\%$ relative lift in one-hop Query CTR and a $3.1\%$ relative lift in two-hop Order CVR, ultimately driving a $4.9\%$ increase in Gross Merchandise Volume (GMV). This strong online performance shows that our Consistency Reward effectively bridges the generative-retrieval gap and turns offline gains into real-world impact. 

\subsection{Ablation Study}

Table~\ref{tab:ablation} shows the incremental contribution of each component on the Industrial dataset. Adding \textbf{Product Retrieval} improves Cons@1 from 0.021 to 0.031 and I\_Hit@1 from 0.102 to 0.122, indicating that grounding generation with catalog retrieval helps reduce invalid queries. Incorporating \textbf{Memory Abstract} brings the significant intermediate gain, boosting Q\_EM from 0.045 to 0.099 and Cons@1 from 0.031 to 0.064, showing that compressed user memory is critical for capturing latent intent beyond raw behavior logs (Details in Appendix~\ref{app:memory_analysis}). Finally, \textbf{RAG Rerank} further raises I\_Hit@1 from 0.172 to 0.261 and Cons@1 from 0.064 to 0.117, confirming that accurate reranking is essential for converting relevant candidates into strong end-to-end retrieval outcomes (More details in Appendix~\ref{app:backbone_ablation}). Although the full model increases p99 latency to 1,973\,ms, this cost is handled in deployment via asynchronous pre-computation (See Appendices~\ref{appendix:online_deployment} for details)

\begin{table}
\centering

\resizebox{\columnwidth}{!}{%
\begin{tabular}{l ccc r}
\toprule
\textbf{Variant}
  & \textbf{Q\_EM} & \textbf{I\_Hit@1} & \textbf{Cons@1}
  & \textbf{Latency} \\
\midrule
Single-Turn RL
  & 0.032 & 0.102 & 0.021 & 136\,ms \\
+\,Product Retrieval
  & 0.045 & 0.122 & 0.031 & 782\,ms \\
+\,Memory Abstract
  & 0.099 & 0.172 & 0.064 & 939\,ms \\
+\,RAG Rerank (\textbf{Full})
  & \textbf{0.177} & \textbf{0.261} & \textbf{0.117} & 1,973\,ms \\
\bottomrule
\end{tabular}%
}
\caption{%
  Ablation Study on the Industrial dataset. Latency denotes p99 online inference time.}
\vspace{-0.9em}
\label{tab:ablation}
\end{table}

%邹正修改
\section{Related Work}
Existing work on query recommendation mainly focuses on explicit user inputs~\cite{mo-etal-2023-convgqr,query_r2, query_re1, jang2024itercqriterativeconversationalquery, search-r1,deepretrieval} while our work focuses on e-commerce query recommendation without user input. Under this setting, previous industrial methods are largely based on co-occurrence matching and semantic retrieval~\cite{itemcf, swing, dssm, qwen3_emb}. Recent studies have also explored generative query recommendation from implicit logs~\cite{aigq,hao2025oxygenrec}. Although effective at the query level, these methods may underemphasize downstream consistency.

\section{Conclusion}
We proposed \textbf{QueryAgent-R1}, a memory-augmented agentic framework for query recommendation in e-commerce search. By grounding query generation in real inventory retrieval and optimizing with a consistency reward, our method bridges the generation-retrieval gap and better aligns recommended queries with downstream business value. Experiments on both offline datasets and online A/B tests show that our QueryAgent-R1 consistently outperforms strong baselines. In future work, we plan to further improve the efficiency of online deployment while maintaining retrieval quality and downstream gains.

\section*{Limitations}
QueryAgent-R1 is currently limited by its online inference efficiency. Due to the relatively long response time of the agentic framework, our production system adopts an asynchronous deployment strategy to satisfy latency requirements. Although this design enables practical large-scale serving, it also increases system complexity and may constrain real-time responsiveness. Future work will focus on reducing online latency and improving inference efficiency to support more flexible deployment.

\section*{Ethical Considerations}
When deploying our QueryAgent-R1, we follow two main ethical principles. 
(1) \textbf{Data privacy protection}: All data used for training and inference are strictly anonymized and de-identified, with no personally identifiable information, platform-specific user IDs, or other sensitive attributes included. The model operates only on sanitized interaction logs, which helps preserve user privacy throughout the entire pipeline. 
(2) \textbf{Content safety}: To reduce the risk of generating inappropriate, sensitive, or harmful suggestions, we explicitly filter sensitive topics during both training and inference. This preventive mechanism helps mitigate potential harms related to unsafe or biased content.

\section*{Acknowledgements}
The authors used generative AI tools only to polish the writing of this manuscript, such as correcting grammar, rephrasing sentences and improving readability. The tools were not used for research idea, methodology design, experimental implementation, data analysis or interpretation of results. 


\bibstyle{acl_nationbib}
\bibliography{main}

\appendix

\section{Experimental Details}
\subsection{Dataset Construction Details}
\label{app:dataset_construction}

We construct a specialized instruction tuning dataset for next query prediction by aligning search relevance annotations with user review content. The curation pipeline is detailed below.

\paragraph{Data Sources and Alignment}
Our corpus builds upon two public datasets:
\begin{itemize}
    \item \textbf{Amazon ESCI Dataset}~\cite{amazon_esci}: This large scale search relevance dataset contains query product pairs annotated as exact (E), substitute (S), complement (C), or irrelevant (I). We retain only US locale records and binarize labels by treating E as positive and I as negative.
    \item \textbf{Amazon Review Dataset}~\cite{amazon_review}: We aggregate reviews from five categories including Magazine Subscriptions, Software, Digital Music, All Beauty, and Amazon Fashion. Each record includes the user ID, ASIN, rating, timestamp, and review text.
\end{itemize}

Given the absence of public click stream logs, we adopt a proxy alignment strategy. An inner join on product ASIN identifies overlapping items between the ESCI and Review datasets. Among approximately 997K unique ESCI ASINs and 1.1M Review ASINs, we find 4,939 common ASINs with a 0.49\% overlap rate. Merging search and review records on these shared ASINs allows us to use review text as a behavioral proxy for post search interaction. After deduplicating on $(user\_id, query, product\_id)$ tuples and keeping only positive relevance samples, we obtain 2,527,263 valid interaction records covering 896,498 unique users.

\paragraph{Behavior Sequence Construction}
Interactions for each user are sorted chronologically and formatted into a unified behavior string:
\begin{equation}
    b_t = \texttt{"search: } q_t \texttt{ [SEP] click: } r_t \texttt{"}
\end{equation}
where $q_t$ is the search query and $r_t$ is the corresponding review text at time step $t$. The full behavior sequence for user $u$ is $\mathcal{S}_u = [b_1, b_2, \dots, b_{|\mathcal{S}_u|}]$. Users with fewer than 10 behaviors are discarded to ensure sufficient contextual history, yielding a final corpus of 17,514 active users.

\paragraph{Next Behavior Prediction Task Formulation}
The training objective follows a next behavior prediction task. Each user sequence $\mathcal{S}_u$ is split into an input context and a prediction target:
\begin{align}
    \mathbf{x}_u &= [b_1, b_2, \dots, b_{|\mathcal{S}_u|-1}] \\
    \mathbf{y}_u &= b_{|\mathcal{S}_u|}
\end{align}
Input behaviors in $\mathbf{x}_u$ are concatenated with newline delimiters to form a readable context, while $\mathbf{y}_u$ serves as the ground truth next behavior for generation.

\paragraph{Instruction Tuning Format}
Each sample is converted into a multi turn conversation format compatible with large language model fine tuning. The system prompt defines the model role as an ecommerce next query prediction engine and enforces strict output constraints: predicted queries must be under 10 words, written in natural search keyword style, and wrapped in \texttt{<next\_query>} XML tags without additional reasoning or markdown formatting. The user prompt provides the chronological behavior history $\mathbf{x}_u$ and instructs the model to predict the single most probable next search query based solely on the provided context.

\paragraph{Dataset Split and Statistics}
A user level stratified split prevents data leakage, allocating 95\% of users to training and 5\% to testing. Final dataset statistics appear in Table~\ref{tab:dataset_stats}. Both splits are stored in Parquet format with structured metadata fields including prompt messages, ability tags, reward model ground truth, and auxiliary information for downstream RLHF evaluation.

\begin{table}[h]
    \centering
    \small
    \begin{tabular}{lcc}
        \toprule
        \textbf{Statistic} & \textbf{Train} & \textbf{Test} \\
        \midrule
        Number of samples & 16,638 & 876 \\
        Avg. behaviors per user & 14.2 & 14.1 \\
        Unique queries & 3,896 & -- \\
        Unique products & 4,939 & -- \\
        Positive label ratio & 100\% & 100\% \\
        \bottomrule
    \end{tabular}
    \caption{Statistics of the constructed next query prediction dataset.}
    \label{tab:dataset_stats}
\end{table}

\subsection{LLM Direct Inference Details}
\label{appendix:LLM_direct_gap_analysis}
As summarized in Table~\ref{tab:full_baseline_appendix}, we observe a consistent disparity between the generative capabilities of LLMs and their retrieval consistency. On Amazon dataset: Gemini-3.1-Pro achieves a state-of-the-art Query Exact Match (Q\_EM) of 0.123, indicating a superior ability to hallucinate or predict potential user search intents. However, its Consistency@1 (Cons@1) plummet to a mere 0.019. 

This phenomenon perfectly illustrates our core premise: accurately predicting a linguistically plausible query does not inherently guarantee the retrieval of relevant, purchasable inventory within a constrained database. This gap is even more pronounced in larger models, suggesting that simply scaling up parameters improves generative fluency but does not necessarily resolve the grounding issue in e-commerce retrieval tasks.

\begin{table*}[t]
\centering
\small
\renewcommand{\arraystretch}{1.2}
\setlength{\tabcolsep}{12pt}

\begin{tabular}{l ccc ccc}
\toprule
& \multicolumn{3}{c}{\textbf{Industrial}} & \multicolumn{3}{c}{\textbf{Amazon}} \\
\cmidrule(lr){2-4} \cmidrule(lr){5-7}
% \multirow{-2}{*}{\textbf{\multicolumn{1}{c}{Method}}} & \textbf{Q\_EM$\uparrow$} & \textbf{I\_Hit@1$\uparrow$} & \textbf{Cons@1$\uparrow$} & \textbf{Q\_EM$\uparrow$} & \textbf{I\_Hit@1$\uparrow$} & \textbf{Cons@1$\uparrow$} \\
\multirow{-2}{*}{\textbf{Method}} & \textbf{Q\_EM$\uparrow$} & \textbf{I\_Hit@1$\uparrow$} & \textbf{Cons@1$\uparrow$} & \textbf{Q\_EM$\uparrow$} & \textbf{I\_Hit@1$\uparrow$} & \textbf{Cons@1$\uparrow$} \\

\midrule

% --- Gemini Family ---
\rowcolor{headergray} \multicolumn{7}{c}{\textbf{Gemini Family}} \\
Gemini-2.0-flash & 0.052 & \cellcolor{better}0.100 & \cellcolor{better}0.036 & \cellcolor{good}0.111 & 0.080 & 0.014 \\
Gemini-2.5-flash & 0.045 & 0.083 & 0.021 & 0.073 & 0.078 & 0.010 \\
Gemini-2.5-pro   & 0.036 & 0.075 & 0.017 & 0.087 & 0.070 & 0.011 \\
Gemini-3-flash & \cellcolor{good}0.053 & 0.070 & 0.020 & 0.074 & 0.054 & 0.009 \\
Gemini-3-pro   & 0.040 & 0.065 & 0.018 & \cellcolor{better}0.115 & \cellcolor{good}0.080 & \cellcolor{good}0.019 \\
Gemini-3.1-pro & \cellcolor{better}0.054 & \cellcolor{good}0.096 & \cellcolor{good}0.029 & \cellcolor{best}0.123 & \cellcolor{better}0.091 & \cellcolor{better}0.021 \\
\midrule

% --- GPT Family ---
\rowcolor{headergray} \multicolumn{7}{c}{\textbf{GPT Family}} \\
GPT-4o-mini & 0.038 & 0.080 & 0.023 & 0.040 & 0.045 & 0.003 \\
GPT-4o      & 0.036 & 0.065 & 0.020 & 0.033 & 0.030 & 0.002 \\
GPT-5       & 0.022 & 0.055 & 0.008 & 0.037 & 0.035 & 0.005 \\
GPT-5.1     & 0.038 & 0.091 & 0.023 & 0.039 & 0.042 & 0.002 \\
\midrule

% --- Qwen Family ---
\rowcolor{headergray} \multicolumn{7}{c}{\textbf{Qwen Family}} \\
Qwen3-max     & 0.034 & 0.060 & 0.021 & 0.035 & 0.047 & 0.001 \\
Qwen3.6-flash & 0.037 & 0.078 & 0.023 & 0.090 & 0.058 & 0.013 \\
Qwen3.6-plus  & 0.047 & 0.084 & 0.028 & 0.096 & 0.058 & 0.015 \\
\midrule

% --- DeepSeek Family ---
\rowcolor{headergray} \multicolumn{7}{c}{\textbf{DeepSeek Family}} \\
DeepSeek-v3.1       & 0.025 & 0.051 & 0.011 & 0.033 & 0.038 & 0.001 \\
DeepSeek-v3.2       & 0.024 & 0.062 & 0.014 & 0.033 & 0.039 & 0.001 \\
DeepSeek-v3.2-exp   & 0.031 & 0.061 & 0.015 & 0.040 & 0.050 & 0.002 \\
DeepSeek-r1         & 0.032 & 0.078 & 0.020 & 0.046 & 0.054 & 0.006 \\
DeepSeek-v4-flash   & 0.049 & 0.064 & 0.023 & 0.070 & \cellcolor{better}0.091 & 0.013 \\
DeepSeek-v4-pro     & 0.034 & 0.070 & 0.018 & 0.082 & \cellcolor{better}0.091 & 0.011 \\
\midrule

% --- Ours ---
\rowcolor{headergray} \multicolumn{7}{c}{\textbf{Ours}} \\
\textbf{QueryAgent-R1} & \cellcolor{best}\textbf{0.177} & \cellcolor{best}\textbf{0.261} & \cellcolor{best}\textbf{0.117} & 0.095 & \cellcolor{best}\textbf{0.144} & \cellcolor{best}\textbf{0.063} \\

\bottomrule
\end{tabular}
\caption{Full baseline results for Gemini, GPT, Qwen, and DeepSeek model families. The top three results in each column are highlighted in \colorbox{best}{first}, \colorbox{better}{second}, and \colorbox{good}{third} places.}
\label{tab:full_baseline_appendix}
\end{table*}

\subsection{Analysis of the Impact of Memory Integration}
\label{app:memory_analysis}

In this section, we provide a detailed empirical analysis of the full baseline results presented in Table~\ref{tab:memory_baseline_appendix}, which evaluates the capability of state-of-the-art Large Language Models (LLMs) to leverage historical context on the Industrial dataset under both with-memory and without-memory experimental settings.

\paragraph{Generality and Quantitative Benefits of Memory.}
Empirical results show that incorporating a memory module yields robust performance gains across diverse model architectures. Among the 22 evaluated models from the Gemini\footnote{\url{https://blog.google/products-and-platforms/products/gemini/}}, GPT\footnote{\url{https://openai.com/api/}}, Qwen\footnote{\url{https://qwen.ai/apiplatform}}, and DeepSeek\footnote{\url{https://huggingface.co/deepseek-ai}} families, 19 (86.4\%) improved on at least one key metric ($Q\_EM$, $I\_Hit@1$, or $Cons@1$). Notably, 10 models (45.5\%), including Gemini-2.0-flash, Gemini-3.1-pro, GPT-5, and most DeepSeek-v3/v4 variants, achieved simultaneous improvements across all three dimensions. These findings confirm that retrieved historical interactions serve as an effective plug-and-play enhancement, enabling models to preserve semantic consistency and make better-informed decisions in complex multi-turn industrial workflows.

\paragraph{The Need for Active Adaptation and Training.}
Despite the general success, we observe that several vanilla LLMs (e.g., GPT-4o-mini, Qwen3.6-plus, and DeepSeek-r1) exhibit minor performance fluctuations or slight degradation on specific metrics when memory is introduced. We attribute this phenomenon to \textit{memory distraction}, as untrained, off-the-shelf models often struggle to distinguish relevant historical cues from retrieve-induced noise or fail to locate critical key-value pairs within dense retrieved slots.

This observation highlights a crucial insight: while zero-shot memory prompt injection is beneficial, co-designing the memory mechanism with targeted model training (e.g., supervised fine-tuning or reinforcement learning) can exponentially magnify the utilization of memory. By explicitly training models on memory-augmentation tasks, we can align the LLM's internal representation with the external memory database. This optimization enables the model to transition from passive context-reading to active memory-filtering and reasoning. Such synergy explains why highly optimized or trained agents, such as our proposed \textbf{QueryAgent-R1}, can seamlessly overcome the distraction barrier and achieve order-of-magnitude higher gains compared to their untrained vanilla counterparts.

\begin{table*}[t]
\centering
\small
\renewcommand{\arraystretch}{1.2}
\setlength{\tabcolsep}{12pt}

\begin{tabular}{l ccc ccc}
\toprule
& \multicolumn{3}{c}{\textbf{Base}} & \multicolumn{3}{c}{\textbf{+Memory}} \\
\cmidrule(lr){2-4} \cmidrule(lr){5-7}
% \multirow{-2}{*}{\textbf{\multicolumn{1}{c}{Method}}} & \textbf{Q\_EM$\uparrow$} & \textbf{I\_Hit@1$\uparrow$} & \textbf{Cons@1$\uparrow$} & \textbf{Q\_EM$\uparrow$} & \textbf{I\_Hit@1$\uparrow$} & \textbf{Cons@1$\uparrow$} \\
\multirow{-2}{*}{\textbf{Method}} & \textbf{Q\_EM$\uparrow$} & \textbf{I\_Hit@1$\uparrow$} & \textbf{Cons@1$\uparrow$} & \textbf{Q\_EM$\uparrow$} & \textbf{I\_Hit@1$\uparrow$} & \textbf{Cons@1$\uparrow$} \\
\midrule

% --- Gemini Family ---
\rowcolor{headergray} \multicolumn{7}{c}{\textbf{Gemini Family}} \\
Gemini-2.0-flash & 0.052 & 0.100 & 0.036 & 0.061\chk & 0.118\chk & 0.045\chk \\
Gemini-2.5-flash & 0.045 & 0.083 & 0.021 & 0.044\cls & 0.094\chk & 0.029\chk \\
Gemini-2.5-pro   & 0.036 & 0.075 & 0.017 & 0.038\chk & 0.070\cls & 0.018\chk \\
Gemini-3-flash   & 0.053 & 0.070 & 0.020 & 0.051\cls & 0.086\chk & 0.030\chk \\
Gemini-3-pro     & 0.040 & 0.065 & 0.018 & -- & -- & -- \\
Gemini-3.1-pro   & 0.054 & 0.096 & 0.029 & 0.056\chk & 0.098\chk & 0.034\chk \\
\midrule

% --- GPT Family ---
\rowcolor{headergray} \multicolumn{7}{c}{\textbf{GPT Family}} \\
GPT-4o-mini           & 0.038 & 0.080 & 0.023 & 0.030\cls & 0.072\cls & 0.017\cls \\
GPT-4o                & 0.036 & 0.065 & 0.020 & 0.022\cls & 0.077\chk & 0.013\cls \\
GPT-5                 & 0.022 & 0.055 & 0.008 & 0.024\chk & 0.060\chk & 0.013\chk \\
GPT-5.1               & 0.038 & 0.091 & 0.023 & -- & -- & -- \\
\midrule

% --- Qwen Family ---
\rowcolor{headergray} \multicolumn{7}{c}{\textbf{Qwen Family}} \\
Qwen3-max           & 0.034 & 0.060 & 0.021 & 0.031\cls & 0.095\chk & 0.018\cls \\
Qwen3.6-flash       & 0.037 & 0.078 & 0.023 & 0.037     & 0.083\chk & 0.023     \\
Qwen3.6-plus        & 0.047 & 0.084 & 0.028 & 0.044\cls & 0.079\cls & 0.020\cls \\
\midrule

% --- DeepSeek Family ---
\rowcolor{headergray} \multicolumn{7}{c}{\textbf{DeepSeek Family}} \\
DeepSeek-v3.1       & 0.025 & 0.051 & 0.011 & 0.037\chk & 0.103\chk & 0.029\chk \\
DeepSeek-v3.2       & 0.024 & 0.062 & 0.014 & 0.032\chk & 0.083\chk & 0.015\chk \\
DeepSeek-v3.2-exp   & 0.031 & 0.061 & 0.015 & 0.039\chk & 0.108\chk & 0.030\chk \\
DeepSeek-r1         & 0.032 & 0.078 & 0.020 & 0.023\cls & 0.070\cls & 0.013\cls \\
DeepSeek-v4-flash   & 0.049 & 0.064 & 0.023 & 0.053\chk & 0.099\chk & 0.040\chk \\
DeepSeek-v4-pro     & 0.034 & 0.070 & 0.018 & 0.048\chk & 0.091\chk & 0.032\chk \\

\bottomrule
\end{tabular}
\caption{Full baseline results for Gemini, GPT, Qwen, and DeepSeek model families based on the Industrial dataset, evaluating the performance impact of incorporating Memory. Green checkmarks (\checkmark) and red crossmarks ($\times$) indicate performance gains and degradation when incorporating Memory, respectively.}
\label{tab:memory_baseline_appendix}
\end{table*}

\subsection{RL Algorithm Comparison}
\label{appendix:rl_comparison}
To further validate the effectiveness of GDPO (Gradient-Decoupled Policy Optimization) in our proposed multi-reward setting, we conduct a comparative study against the standard GRPO (Group Relative Policy Optimization) baseline. The training process involves three distinct reward signals: \textit{Query Hit}, \textit{Item Hit}, and \textit{Grounding Consistency}.

\begin{figure*}[htbp]
     \centering
     % 第一张图：Consistency
     \begin{subfigure}[b]{0.32\textwidth}
         \centering
         \includegraphics[width=\textwidth]{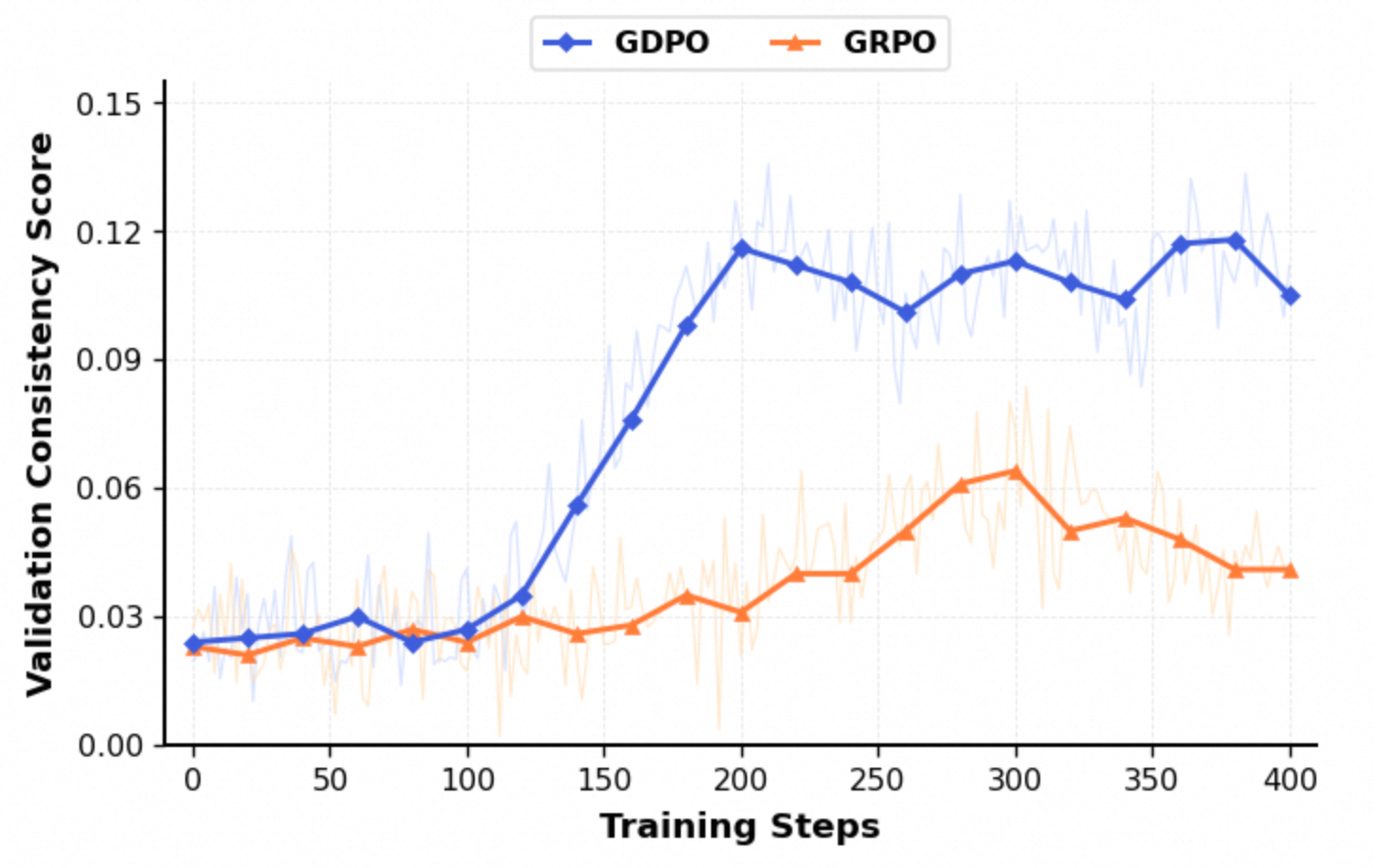} % 请替换为你的文件名
         \caption{Validation Consistency Score}
         \label{fig:val_consistency}
     \end{subfigure}
     \hfill
     % 第二张图：Item Hitrate
     \begin{subfigure}[b]{0.32\textwidth}
         \centering
         \includegraphics[width=\textwidth]{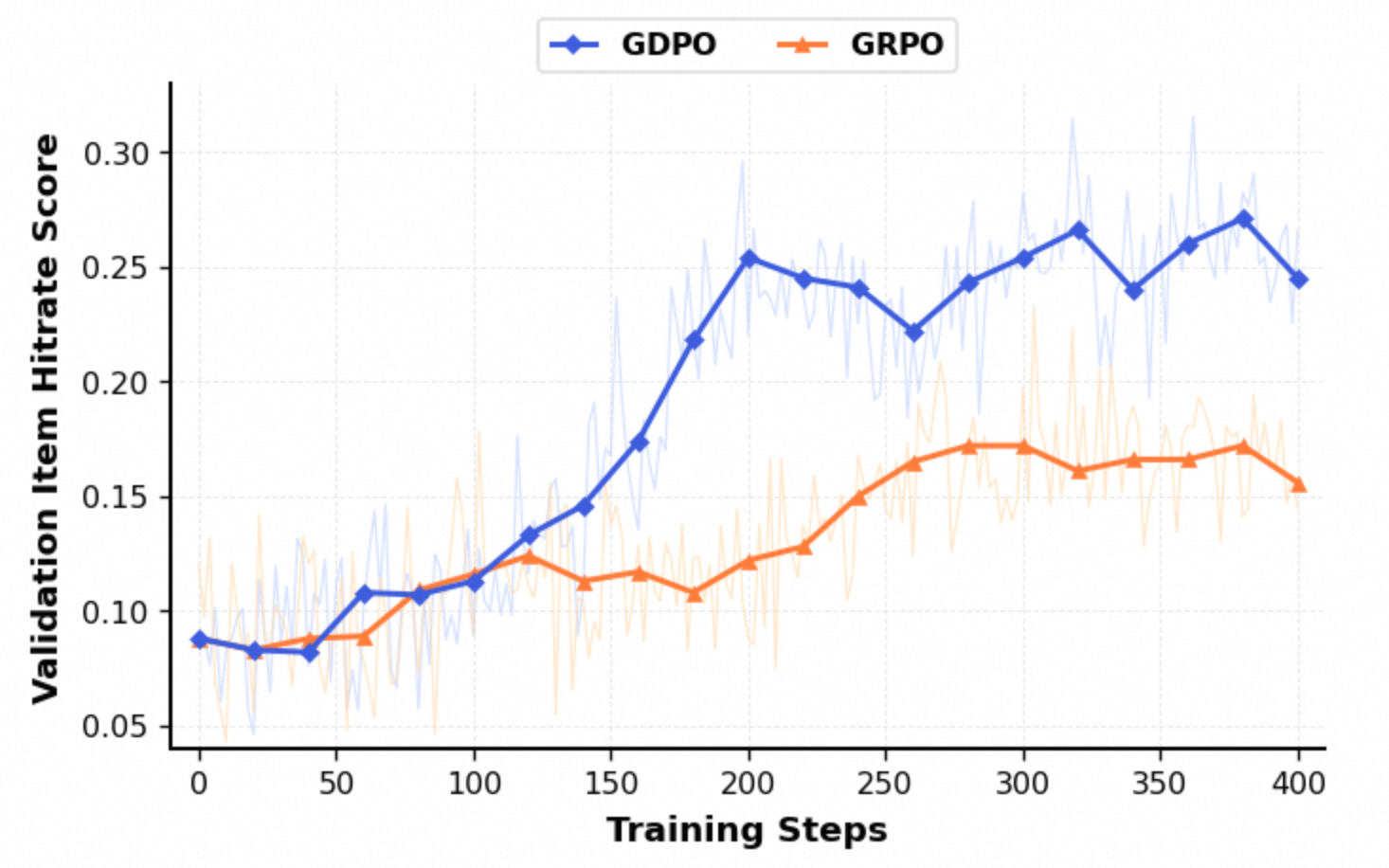} % 请替换为你的文件名
         \caption{Validation Item Hitrate}
         \label{fig:val_item}
     \end{subfigure}
     \hfill
     % 第三张图：Query Hitrate
     \begin{subfigure}[b]{0.32\textwidth}
         \centering
         \includegraphics[width=\textwidth]{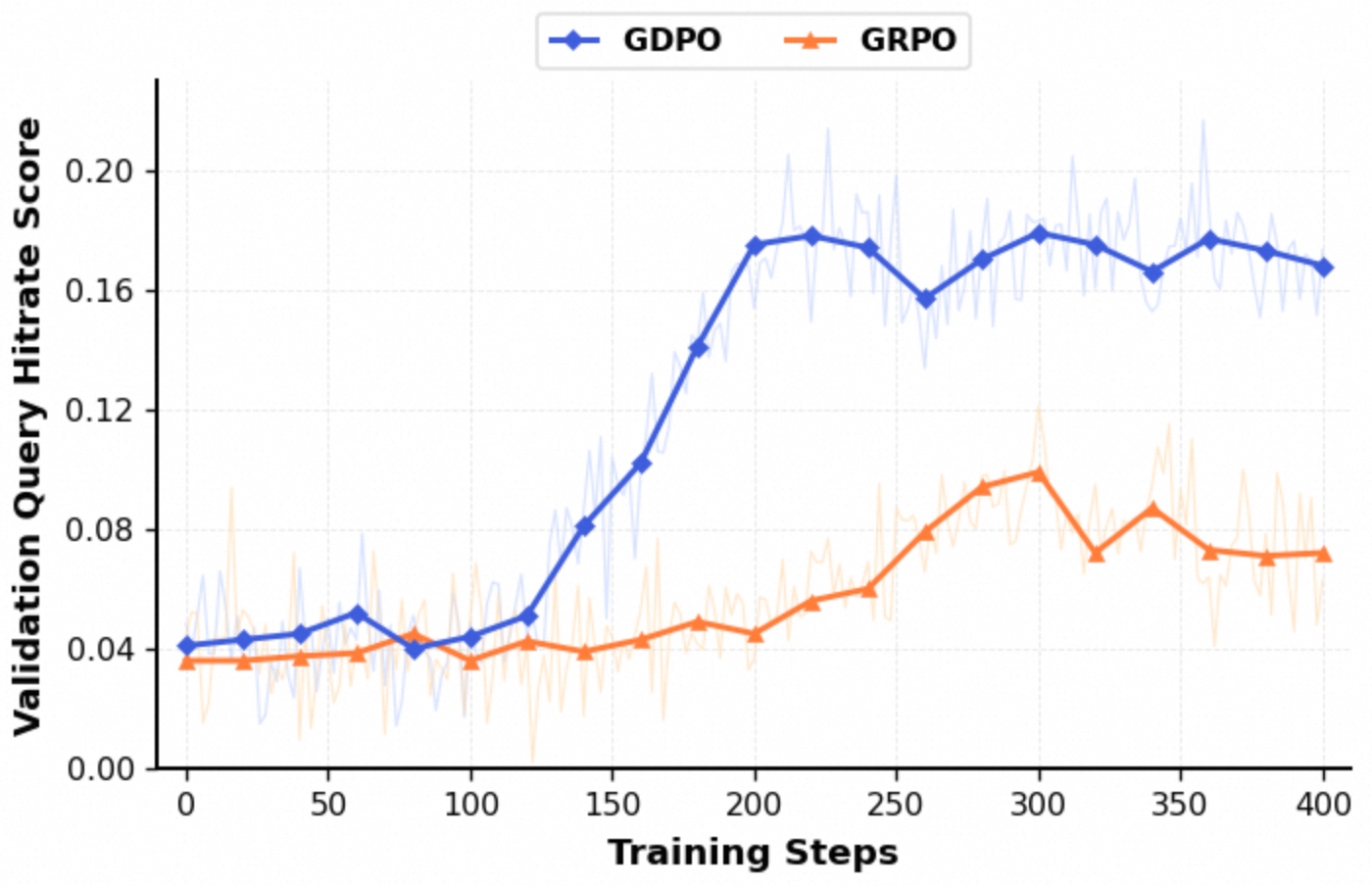} % 请替换为你的文件名
         \caption{Validation Query Hitrate}
         \label{fig:val_query}
     \end{subfigure}
     
     \caption{Training dynamics comparison between GDPO and GRPO. GDPO demonstrates superior convergence and higher asymptotic performance in all three key metrics within the multi-reward optimization framework.}
     \label{fig:rl_curves}
\end{figure*}

\noindent\textbf{Superiority in Multi-Reward Coordination.} 
Figure~\ref{fig:rl_curves} illustrates the validation trajectories of both algorithms over 400 training steps. We observe that while both algorithms perform similarly in the early exploration phase (0-100 steps), GDPO exhibits a significantly steeper learning curve and a higher performance ceiling across all three metrics starting from step 150. 

Specifically, in Figure~\ref{fig:val_consistency}, the \textit{Consistency Score} of GDPO reaches a plateau of approximately 0.12, nearly double that of GRPO ($\approx$0.06). This disparity highlights a critical limitation of GRPO in multi-reward scenarios: gradient interference. GRPO often struggles to balance competing reward signals, leading to sub-optimal policies or "seesaw" effects where one metric improves at the expense of others. In contrast, GDPO effectively decouples the gradients of different objectives, allowing the model to optimize the \textit{Query-Item-Consistency} triad synergistically.

\noindent\textbf{Stability and Convergence.} 
Furthermore, GDPO demonstrates superior training stability. As shown in the \textit{Item Hitrate} and \textit{Query Hitrate} curves (Figure~\ref{fig:val_item} and \ref{fig:val_query}), GDPO maintains a steady upward trajectory with lower variance compared to the highly oscillatory behavior of GRPO. This suggests that our algorithm is more robust to the sparse and noisy nature of reward signals in complex agentic retrieval tasks.

% --- 三图并排布局 ---

\subsection{Ablation Analysis on the Qwen3-4B Backbone}
\label{app:backbone_ablation}

To analyze the scaling behavior and individual contributions of our core algorithmic components, we conduct a cumulative ablation study on the lightweight Qwen3-4B backbone (Table~\ref{tab:ablation_4b}).

\begin{table}[h]
\centering

\renewcommand{\arraystretch}{1.1}
\setlength{\tabcolsep}{4pt} % 1. 压缩列间距，极大收窄表格整体宽度
\footnotesize % 2. 采用标准的学术窄表字号
\resizebox{\columnwidth}{!}{% % 3. 动态自适应包裹，确保100%完美贴合单栏边界
\begin{tabular}{l ccc}
\toprule
\textbf{Variant} & \textbf{Q\_EM} & \textbf{I\_Hit@1} & \textbf{Cons@1} \\
\midrule
\textit{Qwen3-4B Baselines:} & & & \\
~Direct Inference & 0.037 & 0.084 & 0.025 \\
~+\,RAG           & 0.023 & 0.091 & 0.018 \\
~+\,SFT           & 0.052 & 0.054 & 0.029 \\
~+\,RL            & 0.032 & 0.102 & 0.021 \\
~\underline{+\,SFT \& RL} & \underline{0.114} & \underline{0.197} & \underline{0.065} \\
~+\,Rejection Sampling & 0.047 & 0.076 & 0.033 \\
\midrule
\textbf{QueryAgent-R1} (Ours) & \textbf{0.177} & \textbf{0.261} & \textbf{0.117} \\
\bottomrule
\end{tabular}%
}
\vspace{-2mm}
\caption{Ablation study on the Industrial dataset utilizing the \textbf{Qwen3-4B} backbone. Underline indicates the best performance among 4B-based variants; bold denotes the overall best performance.}
\label{tab:ablation_4b}
\end{table}

\paragraph{Analysis of Isolated Components and Rejection Sampling.}
Applying RAG~\cite{rag} directly to the raw 4B backbone degrades $Q\_EM$ (from $0.037$ to $0.023$), showing that small models suffer from context distraction when processing raw retrieved slots. 
For \textbf{+\,Rejection Sampling}, we fine-tune the 4B model on high-quality, successful trajectories generated by our strong teacher model (\textbf{QueryAgent-R1}) and filtered via rejection sampling. While this distillation-style SFT yields stabler formatting ($Q\_EM=0.047$) and logical consistency ($Cons@1=0.033$) compared to standard SFT, its overall reasoning capability is still fundamentally capped by the inherent distribution shift between offline teacher demonstrations and online student rollouts.

\paragraph{The Synergy of SFT \& RL.}
The joint \textbf{+\,SFT \& RL} paradigm achieves the best performance among all 4B variants ($0.114 / 0.197 / 0.065$). Here, SFT provides essential structure by enforcing formatting and task constraints, which drastically narrows the exploration space for subsequent Reinforcement Learning. Guided by this warm-up, online RL successfully optimizes search strategies and avoids policy collapse. 
% Nevertheless, a visible performance gap remains between the optimized 4B agent and the full \textbf{QueryAgent-R1} ($0.177 / 0.261 / 0.117$), highlighting that model scale remains vital for handling highly complex, multi-turn industrial workflows.

\section{Implementation Details}
\label{appendix:implementation_details}
The main policy model (Qwen3-4B) is trained on an $8\times$ H20 GPU cluster using the \textsc{verl} framework~\cite{verl} with Group Reward-Decoupled Normalization Policy Optimization (GDPO) with a group size of $K=8$ and a learning rate of $10^{-5}$ for $3$ epochs over a dataset of $54\text{k}$ training samples. Our retrieval backend executes \textit{Hybrid Retrieval} using FAISS-based dense vectors (Qwen3-Emb-0.6B) and BM25~\cite{bm25} sparse matching, combined with a Qwen3-Rerank-0.6B module to score query-item-behavior triplets.

\smallskip
\noindent\textbf{Offline Training.} 
To support high-concurrency tool-calling during RL rollouts, we encapsulate the RAG Search Tool as a distributed FastAPI sandbox powered by Ray. The retrieval backend executes \textit{Hybrid Retrieval} using FAISS-based dense vectors (Qwen3-Emb-0.6B) and BM25~\cite{bm25} sparse matching, followed by a Qwen3-Rerank-0.6B module to precisely score query-item-behavior triplets. To prevent Sandbox overload during massive policy exploration, we implement a token-bucket rate limiter. The main policy model (Qwen3-4B) is trained on 8$\times$H20 GPUs using standard GDPO (group size $K{=}8$, learning rate $10^{-5}$) for 3 epochs over 54k samples.

\smallskip
\noindent\textbf{Online Serving.} 
Memory Abstraction profiles are pre-computed via daily batch jobs. The online agent runs on a single A10 GPU using vLLM~\cite{vllm}. As established in Section~4.3, the ${\sim}$2,000\,ms p99 latency is fully absorbed by our asynchronous deployment. Furthermore, because the generated queries are strictly grounded in inventory, we can bypass the expensive downstream ranking stages of the legacy pipeline, maintaining strict computational cost parity.

\section{Tool-Use Infrastructure}
\label{appendix:sandbox}

\subsection{Resilient Distributed Tool-Calling Infrastructure.} 
During parallel RL rollouts, the high frequency of environmental feedback requests presents severe concurrency challenges. To scale the RAG Search Tool safely, we design a cluster-wide singleton execution framework using Ray. Rather than initializing independent rate limiters per agent thread, which leads to aggregate bottleneck failures, we register named, detached Ray actors cluster-wide. Specifically, a \textit{Centralized Token-Bucket Limiter} implements a thread-safe token-bucket algorithm using asynchronous semaphore orchestration across dedicated \texttt{acquire} and \texttt{release} concurrency groups. Simultaneously, a \textit{Global Worker Pool} acts as a cluster-wide actor singleton that schedules non-blocking asynchronous calls via \texttt{await} execution to optimize throughput. Furthermore, we implement a defensive transaction layer within the execution workers: if downstream APIs experience transient failure under peak load, a fault-tolerant HTTP pipeline automatically triggers up to $10$ retries with dynamic backoff delay scaling ($t_{\text{delay}} = \text{Initial Delay} \times \text{Attempt}$). This robust systems-level design eliminates environment-induced rollout crashes, securing deterministic and high-throughput RL training trajectories.

\subsection{User Memory Sandbox Architecture}

To serve user memory profiles during RL rollouts, we run a dedicated memory retrieval service encapsulated as a Ray actor allocated with $1$ CPU and $0$ GPUs. This minimal allocation prevents resource contention with the primary policy models during parallel rollout generation. 

\paragraph{Distributed Deployment and API Hosting}
Inside the Ray actor, an asynchronous FastAPI server hosted via Uvicorn manages incoming HTTP requests. The server runs on a lightweight event loop to handle concurrent profile fetches from the rollout workers without blocking the environment simulation. The service logs processing latencies and file access paths to monitor system performance under high query throughput.

\paragraph{Identifier Resolution Pipeline}
Because online user identifiers (\texttt{user\_id}) often contain characters incompatible with local file systems (such as slashes), the sandbox implements a deterministic matching sequence to locate the corresponding profile JSON files. Upon receiving a request, the resolver first searches for a file named after the raw \texttt{user\_id}. If not found, it falls back to replacing forward slashes (\texttt{/}) with hyphens (\texttt{-}). If a match still fails, it applies URL percent-encoding to the identifier. This fallback sequence prevents runtime exceptions and request failures caused by encoding discrepancies between the database and the file system.

\paragraph{Memory Formatting and Representation}
Directly injecting raw, nested JSON profiles into the LLM context increases input prompt lengths and can lead to attention drift. To optimize token usage, the sandbox formats the parsed JSON data into a structured textual representation. The formatter extracts specific fields including short-term transactional intents, style tags, demographic identities, and price segments. To represent user interests compactly, the key-value pairs of the user's interest graph are sorted in descending order by their numerical weights and formatted as a comma-separated list. These extracted attributes are then compiled into a plain-text template before being returned to the policy agent.

\subsection{Hybrid Search and Reranking Sandbox Architecture}

To execute real-time product retrieval and preference-aware ranking during parallel reinforcement learning rollouts, we deploy a hybrid search sandbox. This infrastructure coordinates dense and sparse retrieval engines, merges candidate spaces, and refines results through a generative sequence classifier.

\paragraph{Distributed Resource Allocation and Web Hosting}
The retrieval and reranking engine is encapsulated as a high-concurrency Ray actor allocated with $16$ CPUs and $8$ GPUs. This substantial compute allocation allows the sandbox to absorb high-throughput batch queries generated by the policy rollouts without introducing pipeline stalls. Inside the actor, an asynchronous FastAPI web service managed by Uvicorn processes incoming requests. The sandbox hosts two primary network architectures in half-precision (\texttt{FP16}): a dense bi-encoder transformer for vector search and an autoregressive decoder model for personalized ranking. Both models use automatic device mapping to optimize memory utilization across the available GPU nodes.

\paragraph{Dense-Sparse Hybrid Retrieval Pipeline}
The retrieval stage operates by executing complementary dense and sparse search pipelines to maximize query-product recall. 
For dense retrieval, incoming query sequences are concatenated with task instructions and passed through the bi-encoder model. The system extracts query embeddings by applying last-token pooling over the final hidden states. These representations are normalized using $L_2$ distance metrics and searched against a pre-compiled index using FAISS~\cite{faiss}. 
For sparse retrieval, the sandbox runs a tokenized, lowercase keyword search against a pre-computed BM25 index built over the entire product document corpus. 
The candidates returned from the dense FAISS lookup and the sparse BM25 search are merged and deduplicated. This hybrid approach ensures that the candidate pool captures both high-level semantic contexts and precise keyword-level matches.

\paragraph{Generative Reranking and Logit Calibration}
Following candidate retrieval, the candidate pool is pruned and reordered based on a localized user transaction history database. The reranker maps the incoming user identifier to their historical interaction records (e.g., clicks and browse behavior) to construct a detailed ranking prompt. This prompt combines the task instruction, the user's interaction logs, the search query, and the candidate product text. 
Rather than decoding tokens autoregressively, the causal language model functions as a binary classifier. The system isolates the raw output logits of the next-token prediction at the exact vocabulary coordinates of the target tokens ``yes'' and ``no''. 
To prevent probability distortion, these raw logits are isolated and calibrated using a log-softmax function. The exponentiated score of the positive class token (``yes'') represents the final relevance score. The candidate products are then sorted in descending order based on this score, and the top-ranked products are returned to the policy agent.

\section{Online Deployment Details}
\label{appendix:online_deployment}

The post-trained Qwen3-4B agent is deployed online using the vLLM inference engine~\cite{vllm} hosted on a single NVIDIA L20 GPU. Under this configuration, the single-node serving throughput is approximately $10$ queries per second (QPS). Due to these computational throughput constraints, we route a controlled $1\%$ subset of live production traffic to the agent pipeline.

\subsection{Near-Line Latency Masking}
The end-to-end inference latency of the agent is approximately $2.0$ seconds (p99), which exceeds the strict sub-hundred-millisecond budget required for direct synchronous homepage rendering. To resolve this without degrading the user experience, we implement a \textit{near-line} (asynchronous) serving strategy leveraging user navigation paths. The agent's query generation is preemptively triggered the moment a user clicks a product to enter its detail page. Because the user's dwell time on the detail page before returning to the homepage typically spans $5$ to $10$ seconds, the $2.0$-second execution latency of the agent is fully masked behind the client-side navigation delay. This ensures that the personalized query recommendation is already computed and cached by the time the user returns to the homepage.

\subsection{Online Tool and Memory Infrastructure}
User memory profiles are constructed via offline batch pipelines and indexed in an online Redis cache to enable low-latency retrieval. Once a user accumulates a threshold of real-time behavioral signals, the memory updater runs an incremental writeback to update the corresponding Redis entry. The product retrieval tool used by the agent during online inference remains identical to the hybrid dense-sparse search service described in the training phase, ensuring tool-calling consistency between offline optimization and online execution.

\section{Prompt}
\subsection{Training \& Inference Prompt}
\label{appendix:training_prompt}
In this section, we present the prompt used for training and inference, as shown in Figure~\ref{fig:system_prompt_judge_style} and Figure~\ref{fig:user_prompt_judge_style}. It instructs the agent to predict the next search query from the user’s chronological behavior sequence, while mandating memory retrieval and optional search tool use when additional product context is needed. The prompt also enforces concise natural-language outputs and a strict XML-based response format, ensuring consistent query generation and tool-calling behavior.

\begin{figure*}[htbp]
\centering
\begin{judgepromptbox}{Query Agent Prompt Template (SYSTEM\_ROLE)}
\begin{flushleft}
\ttfamily\small
\#\#\# Role Definition \\
You are an E-commerce Query Prediction \& Search Agent. \\
Your task is to analyze the user's chronological real-time behavior logs, determine if additional information is needed via search, and predict the single most likely next search query they will type.

\vspace{1em}
\#\#\# Generation Rules \\
1. Context Awareness: Derive the next query strictly based on the progression of the provided behavior logs. \\
2. User Memory Retrieval (MANDATORY): \\
~~~- Requirement: Before predicting any query, you MUST first retrieve the user's long-term profile and preferences to understand their intent. \\
~~~- Execution: Call the memory tool using the user's ID found in the context. \\
~~~- Syntax: To call the memory tool: \textless{}tool\_call\textgreater{} user\_id \textless{}/tool\_call\textgreater{}. \\
~~~- Integration: Use the retrieved profile (price tier, brand preference, style tags) to refine and personalize the predicted query. \\
3. Tool Usage \& Refinement: \\
~~~- Execution: Verify if you have enough product info context. If info is missing or context is complex, you MUST call Tool(search). \\
~~~- Syntax: To call a tool: \textless{}tool\_call\textgreater{} query + user\_id \textless{}/tool\_call\textgreater{}. \\
~~~- Result Validation: If search results are available, evaluate whether they cover the products user clicked subsequently. \\
~~~- Query Adjustment: Adjust the predicted query to bridge any coverage gaps identified between search results and user behavior. \\
~~~- Reasoning: Use \textless{}reason\textgreater{}...\textless{}/reason\textgreater{} to delineate this thought process and tool check. \\
4. Conciseness: The generated query must be under 10 words. \\
5. Natural Language: The query should sound like a natural user search input (keywords + modifiers), not a full sentence. \\
6. No Explanations: Do not output any extra text outside the defined XML tags.

\vspace{1em}
\#\#\# Output Format Rules \\
\begin{verbatim}
1. Strict XML: Output must be wrapped in <next_query> tags.
2. No Markdown: Do not use markdown code blocks (e.g., ```xml).
3. Structure:
   ~~~[Reasoning steps and tool check]
   ~~~<next_query>
   ~~~[predicted query string]
   ~~~</next_query>
\end{verbatim}

\vspace{1em}
\# Tools \\
You may call one or more functions to assist with the user query. \\
You are provided with function signatures within \textless{}tools\textgreater{}\textless{}/tools\textgreater{} XML tags: \\
\textless{}tools\textgreater{} \\
\{"type": "function", "function": \{"name": "search", "description": "Searches for relevant products based on the given queries.", "parameters": \{"type": "object", "properties": \{"query\_list": \{"type": "array", "description": "List of search queries to retrieve related products."\}, "user\_id": \{"type": "string", "description": "The unique identifier of the user to fetch history behaviors for."\}\}, "required": ["query\_list", "user\_id"]\}\}\} \\
\{"type": "function", "function": \{"name": "memory", "description": "Retrieves the memory logs or profile for a specific user.", "parameters": \{"type": "object", "properties": \{"user\_id": \{"type": "string", "description": "The unique identifier of the user to fetch memory for."\}\}, "required": ["user\_id"]\}\}\} \\
\textless{}/tools\textgreater{}

\vspace{0.5em}
For each function call, return a json object with function name and arguments within \textless{}tool\_call\textgreater{}\textless{}/tool\_call\textgreater{} XML tags: \\
\textless{}tool\_call\textgreater{} \\
\{"name": \textless{}function-name\textgreater{}, "arguments": \textless{}args-json-object\textgreater{}\} \\
\textless{}/tool\_call\textgreater{}
\end{flushleft}
\end{judgepromptbox}
\caption{System prompt template used to instruct the Query Prediction and Search Agent, enforcing multi-tool calling protocols and target formatting constraints.}
\label{fig:system_prompt_judge_style}
\end{figure*}

\begin{figure*}[htbp]
\centering
\begin{judgepromptbox}{Query Agent Prompt Template (USER\_ROLE)}
\begin{flushleft}
\ttfamily\small
\#\#\# Input Data \\
user\_id: \{user\_id\}

\vspace{1em}
Below is the chronological list of the user's real-time search and interaction behaviors:

\vspace{1.2em}
\{realtime\_behavior\_sequence\}

\vspace{1.5em}
\#\#\# Instruction \\
Based strictly on the behavior sequence above, generate the single most probable next search query. \\
Output the result immediately in the required XML format.
\end{flushleft}
\begin{flushleft}
\end{flushleft}
\end{judgepromptbox}
\caption{User prompt template representing the input structure, dynamically populating real-time behavior sequences and localized user tags.}
\label{fig:user_prompt_judge_style}
\end{figure*}

\subsection{Memory Abstract Prompt}
In this section, we present the prompt for memory abstraction. As shown in Figure~\ref{fig:memory_abstract_prompt}, this prompt is used to compress raw user behavioral logs into a structured memory profile for future retrieval. The prompt also enforces a strict JSON output schema, enabling consistent memory updates for downstream agent reasoning.

\begin{figure*}[htbp]
\centering
\begin{judgepromptbox}{Memory Abstract Prompt Template (SYSTEM\_MESSAGES)}
\begin{flushleft}
\ttfamily\small
You are an expert \textbf{User Memory Manager} for an e-commerce AI Agent. Your goal is to analyze raw user behavioral logs and distill them into a structured, abstract user profile for future retrieval.

\vspace{1em}
\textbf{Input Format:} \\
You will receive a list of sequential actions performed by a specific user (\texttt{user\_id}). The data includes item descriptions, categories, timestamps, and prices.

\vspace{1em}
\textbf{Processing Logic (Abstract \& Layering):} \\
Do not simply memorize the list of items. Instead, analyze the data to generate a memory update based on the following layers:

\vspace{0.5em}
1. \textbf{Level 1: Short-term Intent (The "Now")} \\
~~* Analyze recent timestamps and item clusters. \\
~~* Determine the immediate goal (e.g., "Shopping for Christmas decorations," "Stocking up on household essentials," "Looking for jewelry").

\vspace{0.5em}
2. \textbf{Level 2: Abstract Preferences (The "Taste")} \\
~~* Extract stylistic keywords from `action\_beh` (e.g., "French Retro", "Geometric"). \\
~~* Infer price sensitivity based on `action\_price`. \\
~~* Identify preferred brands or shops.

\vspace{0.5em}
3. \textbf{Level 3: Long-term Profile (The "Identity")} \\
~~* Infer high-level categories (e.g., "Home Owner," "Pet Owner" based on 'Pet best supplies'). \\
~~* Summarize key category interests (e.g., 'Jewellery', 'Window Treatment').

\vspace{1em}
\textbf{Output Schema (JSON):} \\
Please output the analysis in the following JSON format strictly: \\
\{ \\
~~"user\_id": "string", \\
~~"memory\_update\_timestamp": "string", \\
~~"short\_term\_intent": "Summary string", \\
~~"style\_tags": ["tag1", "tag2"], \\
~~"price\_segment": "Low/Medium/High", \\
~~"potential\_identities": ["identity1", "identity2"], \\
~~"interest\_graph": \{ \\
~~~~"category\_name": "score (1-10)" \\
~~\}, \\
~~"summary\_text": "A concise natural language summary of this user's recent behavior for the Agent to read." \\
\}
\end{flushleft}
\end{judgepromptbox}
\caption{System prompt template for the User Memory Manager, responsible for abstracting sequential, fine-grained interaction logs into layered, structured JSON user profiles.}
\label{fig:memory_abstract_prompt}
\end{figure*}

\end{document}